\documentclass[journal]{IEEEtran}
\usepackage{arydshln}

\usepackage{amsmath,amsfonts}
\usepackage{algorithm}
\usepackage[rightComments=true]{algpseudocodex}

\usepackage{svg}
\usepackage{array}
\usepackage[caption=false,font=normalsize,labelfont=sf,textfont=sf]{subfig}
\usepackage{textcomp}
\usepackage{stfloats}
\usepackage{url}
\usepackage{verbatim}
\usepackage{graphicx}
\usepackage{cite}
\usepackage{hyperref}
\hypersetup{
    colorlinks=true,
    linkcolor=blue,
    anchorcolor=red,
    filecolor=red,
    menucolor=ref,
    runcolor=red,
    urlcolor=blue,
    citecolor=black,
}
\usepackage{threeparttable}  

\usepackage{booktabs}
\usepackage{multirow}
\usepackage{makecell}

\newcommand*{\footnotemarkcolor}{black}
\makeatletter
\renewcommand*{\@makefnmark}{\hbox{\@textsuperscript{%
  \color{\footnotemarkcolor}\normalfont\@thefnmark}}}
\makeatother

\begin{document}

\title{Distributed Evolution Strategies with Multi-Level Learning for Large-Scale Black-Box Optimization}

\author{Qiqi Duan, Chang Shao, Guochen Zhou, Minghan Zhang, Qi Zhao, Yuhui Shi~\IEEEmembership{Fellow,~IEEE}
	\thanks{This work is supported by the Guangdong Basic and Applied Basic Research Foundation under Grants No. 2024A1515012241 and 2021A1515110024, the Shenzhen Fundamental Research Program under Grant No. JCYJ20200109141235597, and the Program for Guangdong Introducing Innovative and Entrepreneurial Teams under Grant No. 2017ZT07X386.}
	\thanks{Qiqi Duan is with Harbin Institute of Technology, Harbin, China and Southern University of Science and Technology, Shenzhen, China. (e-mail: 11749325@mail.sustech.edu.cn)}
	\thanks{Chang Shao is with Australian Artificial Intelligence Institute, University of Technology Sydney, Sydney, Australia.}
	\thanks{Minghan Zhang is with University of Warwick, Coventry, UK.}
	\thanks{Guochen Zhou, Qi Zhao, and Yuhui Shi are with Department of Computer Science and Engineering, Southern University of Science and Technology, Shenzhen, China. (e-mail: shiyh@sustech.edu.cn).}
	\thanks{Manuscript received X X, 2023.}
}

\markboth{IEEE TRANSACTIONS ON PARALLEL AND DISTRIBUTED SYSTEMS, VOL. 35, NO. 11, NOVEMBER 2024}%
{Duan \MakeLowercase{\textit{et al.}}:}

\maketitle

\begin{abstract}
	In the post-Moore era, main performance gains of black-box optimizers are increasingly depending on parallelism, especially for large-scale optimization (LSO). Here we propose to parallelize the well-established covariance matrix adaptation evolution strategy (CMA-ES) and in particular its one latest LSO variant called limited-memory CMA-ES (LM-CMA). To achieve efficiency while approximating its powerful invariance property, we present a multilevel learning-based meta-framework for distributed LM-CMA. Owing to its hierarchically organized structure, Meta-ES is well-suited to implement our distributed meta-framework, wherein the outer-ES controls strategy parameters while all parallel inner-ESs run the serial LM-CMA with different settings. For the distribution mean update of the outer-ES, both the elitist and multi-recombination strategy are used in parallel to avoid stagnation and regression, respectively. To exploit spatiotemporal information, the global step-size adaptation combines Meta-ES with the parallel cumulative step-size adaptation. After each isolation time, our meta-framework employs both the structure and parameter learning strategy to combine aligned evolution paths for CMA reconstruction. Experiments on a set of large-scale benchmarking functions with memory-intensive evaluations, arguably reflecting many data-driven optimization problems, validate the benefits (e.g., effectiveness w.r.t. solution quality, and adaptability w.r.t. second-order learning) and costs of our meta-framework.
\end{abstract}


\section{Introduction}

\IEEEPARstart{A}{s} both Moore's law \cite{moore2006cramming,moore1998cramming} and Dennard's law \cite{bohr200730,liu2020retrospective} come to end \cite{doi:10.1098/rsta.2019.0061}, main gains in computing performance will come increasingly from the top of the computing stack (i.e., algorithm developing, software engineering, and hardware streamlining) rather than the bottom (semiconductor technology) \cite{doi:10.1126/science.aam9744}. Refer to e.g., the latest \textit{Science} review \cite{doi:10.1126/science.aam9744} or the \textit{Turing} lecture \cite{10.1145/3282307} for an introduction to the modern computing stack.
As recently emphasized by two Turing-Award winners (i.e., Hennessy and Patterson), \textit{“multicore \cite{hill2008amdahl} shifted responsibility for identifying parallelism and deciding how to exploit it to the programmer...”} \cite{10.1145/3282307}. To follow this multi/many-core trend, in this paper we explore the parallelism of evolutionary algorithms (EAs \cite{fogel2006evolutionary}), since intuitively their population-based (random) sampling strategies \cite{10.7551/mitpress/1090.001.0001,back1997evolutionary} are well-suited for massive parallelism \cite{eiben2015evolutionary,10.1162/evco.1993.1.1.1,doi:10.1126/science.aax4953,10.1162/evco_a_00213}.

Specifically, we consider the derandomized evolution strategy with covariance matrix adaptation (CMA-ES \cite{hansen2023cma,10.1162/106365601750190398,hansen1996adapting}) and in particular its one latest variant called limited-memory CMA (LM-CMA \cite{10.1162/EVCO_a_00168,loshchilov2018large}) for large-scale black-box optimization (BBO). As stated in the popular \textit{Nature} review \cite{eiben2015evolutionary}, \textit{“CMA-ES is widely regarded as (one of) the state of the art in numerical (black-box) optimization”} with competitive performance on many benchmarking functions \cite{kern2004learning,10.1145/1830761.1830790,hansen2022anytime,VARELAS2020106737,10.1145/3447929.3447930} and challenging applications (such as \cite{doi:10.1080/13588265.2017.1331493,doi:10.1126/science.aal5054,akrour2014programming,lange2023discovering,doi:10.1126/scirobotics.aat3536,doi:10.1126/sciadv.abq0279}, just to name a few). Although typically its absolute runtime is of minor relevance \cite{10.1162/106365601750190398} for low-dimensional (e.g., $\le$ 50) cases, it cannot be ignored in the distributed computing context for large-scale (e.g., $\ge$ 1000) optimization (LSO) because of its (at least) quadratic complexity w.r.t. each sampling. Since its limited-memory variant (LM-CMA) can fit better for memory hierarchy and distributed communication, our aim is to extend LM-CMA to the modern cloud/clustering computing environment for LSO, in order to show effectiveness \cite{10.1145/121973.121975} while approximating the attractive \textit{invariance} property of CMA-ES \cite{JMLR:v18:14-467} as much as possible.

In their seminal paper \cite{10.1162/106365601750190398}, Hansen and Ostermeier pointed out four fundamental demands for ESs: adaptation, performance, invariance \cite{10.1007/3-540-45356-3_35}, and stationarity (i.e., unbiasedness under random/neutral selection). Clearly, these demands\footnote{As was previously stated in the classical ES review by Beyer and Schwefel \cite{beyer2002evolution}, if some principle is deviated when we design the ES-based variant, the designed optimizer often needs to be widely tested.} are also highly desirable for any distributed ES (DES), in order to obtain efficiency and generalization/transferability \cite{Hansen2014}.
To meet these demands in the distributed computing environment, we adopt the multilevel learning perspective for evolution (recently published in \textit{PNAS} \cite{doi:10.1073/pnas.2120037119,doi:10.1073/pnas.2120042119}) to model and design the efficient DES framework, wherein Meta-ES \cite{10.1162/evco.2008.16.2.151,10.1145/2460239.2460242} and LM-CMA could be naturally combined together to enjoy the best of both worlds.

With the rise of deep models and big data, currently there are increasing needs to optimize high-dimensional objective functions. Among them, a number of black-box scenarios from e.g., non-differentiable simulations and non-convex mathematical models urgently require black-box optimizers to obtain satisfactory performance in a reasonable runtime. Given the fact that almost all of serial black-box optimizers easily suffer from the \textit{curse-of-dimensionality} issue, parallelism is a very natural way to scale up them to large-scale optimization, which is the focus of our paper. Three main contributions of our paper to DES for large-scale BBO are presented as below:

1) We first analyze the parallelism opportunities and challenges of two different ES families (i.e., CMA-ES and Meta-ES) under two common models (i.e., coordinator-worker and island). See Section \ref{sec:related_works} for details. We argue that multilevel learning for biological evolution (MLE) is a natural way to hierarchically combine LM-CMA with Meta-ES under distributed computing (Section \ref{sec:multilevel_meta_framework}).

2) Inspired by MLE \cite{doi:10.1073/pnas.2120037119}, we propose a multilevel learning-based meta-framework for DES to exploit \textit{spatio-temporal} information (if available) on-the-fly to accelerate convergence while maintaining meta-population diversity. Within it, four following key design choices for DES are made reasonably, in order to balance search efficiency (w.r.t. convergence rate, absolute runtime, and diversity) and extra computing cost brought by distributed enhancements (e.g., distributed scheduling, load balancing, fault-tolerance, and data exchanges over the network, etc.) \cite{10.1145/3127479.3128601}.

\begin{itemize}
	\item Owing to its hierarchically organized structure, Meta-ES \cite{10.1145/3377929.3389938} is well-suited to implement the multilevel meta-framework for DES (Section \ref{subsec:hierarchical_organization}).
	\item At the outer-ES level, both the elitist and multi-recombination strategy are used in a parallel fashion in order to alleviate both the stagnation and regression \cite{ARNOLD200618} issue (Section \ref{subsec:update_outer_es_mean}).
	\item The global step-size self-adaptation for DES \cite{rudolph1992correlated} combines the Meta-ES strategy with the well-known cumulative step-size adaptation (CSA), in order to exploit both the \textit{spatial} and \textit{temporal} (non-local) information \cite{JMLR:v13:bergstra12a,10.1145/3067695.3076059} (Section \ref{subsec:adapt_global_step_size}).
	\item To keep a sensible trade-off between efficiency and stability for DES, we extend the collective learning strategy \cite{10.1007/978-3-642-73953-8_8} to distributed computing from two aspects: i) aggregation learning of evolution paths; ii) structure learning of CMA reconstruction (Section \ref{subsec:collective_learning_cma}).
\end{itemize}

3) To validate the effectiveness \cite{10.5555/2831090.2831104} of our proposed meta-framework, we conduct numerical experiments on a large set of large-scale BBO functions. Experimental results show the benefits (and also cost) of our proposed DES meta-framework for large-scale BBO (Section \ref{sec:exp}).

\section{Related Works}\label{sec:related_works}

\setlength{\arrayrulewidth}{0.8pt}

\begin{table*}[!t]
	\centering
	\caption{A Summary of Different ES Versions (w.r.t. Assumption, Complexity, Model Expressability, and Challenge for LSO)}\label{tab:difference_with_exsited_work}
	\resizebox{0.98\textwidth}{!}{%
		\begin{tabular}{c|c|c|c|c}
			\toprule[1.5pt]
			\textbf{ES Version}                                                             & \textbf{Assumption}        & \textbf{Complexity} & \textbf{Model Expressability} & \textbf{Challenge for LSO}               \\
			\midrule[1.0pt]
			DiBB\cite{10.1145/3512290.3528764}                                              & partial separability       & $o(p^2)$            & high                          & weakness on non-separable landscapes     \\
			\midrule[1.0pt]
			piES\cite{jostins2010reverse}                                                   & non-separability           & $o(n)$              & low                           & adaptation of only individual step-sizes \\
			\midrule[1.0pt]
			asynchronous NES\cite{10.1145/2463372.2463424}                                  & non-separability           & $o(n^3)$            & very high                     & excessive requirements in CPU memory     \\
			\midrule[1.0pt]
			asynchronous CMAES\cite{10.1007/978-3-642-37192-9_52}                           & non-separability           & $o(n^2)$            & high                          & difficulty in analyzing procedures       \\
			\midrule[1.0pt]
			distributed ES\cite{10.1007/BFb0029754}                                         & island                     & $o(n)$              & low                           & adaptation of only the global step-size  \\
			\cdashline{1-5}[3pt/1pt]
			\rule{0pt}{1.1\normalbaselineskip}
			VKDCMA\cite{10.1007/978-3-319-45823-6_1}                                        & dynamic low-rank           & $o(nm)$             & modest                        & no parallelism                           \\
			\midrule[1.0pt]
			VDCMA\cite{10.1145/2908812.2908863}, MMES\cite{he2020mmes}                      & fixed low-rank             & $o(n\log(n))$       & modest                        & no parallelism                           \\
			\midrule[1.0pt]
			R1ES\cite{li2017simple}, R1NES\cite{10.1145/2001576.2001692}                    & predominated direction     & $o(n)$              & low                           & no parallelism                           \\
			\midrule[1.0pt]
			DDCMA\cite{li2017simple}                                                        & separability \& invariance & $o(n^2)$            & high                          & no parallelism                           \\
			\midrule[1.0pt]
			SEPCMAES\cite{10.1007/978-3-540-87700-4_30}, SNES\cite{10.1145/2001576.2001692} & separability               & $o(n)$              & low                           & no parallelism                           \\
			\midrule[1.0pt]
			MAES\cite{10.1145/3377929.3389870}, CCMAES\cite{NIPS2016_289dff07}              & invariance                 & $o(n^2)$            & high                          & no parallelism                           \\
			\bottomrule[1.5pt]
		\end{tabular}%
	}
	\begin{tablenotes}
		\item $>$ : $n$ is the dimensionality of problem to be optimized.
		\item $>$ : $m~ (<= n)$ is the total number of evolution paths used to reconstruct the covariance matrix (typically $m = \log(n)$).
		\item $>$ : $p~ (<= n)$ is the maximal dimensionality of all subspaces after decision space decomposition.
	\end{tablenotes}
\end{table*}%

In this section, we only review parallel/distributed versions and large-scale variants of ES, since there have been some well-written reviews for ES (e.g., \cite{beyer2002evolution,kern2004learning,Hansen2006,back2013contemporary,Rudolph12,Hansen2015}) up to now.
We also analyze the opportunities and challenges of two different ES families (i.e., CMA-ES and Meta-ES) under two parallelism models (i.e., coordinator-worker and island).

\subsection{Parallel/Distributed Evolution Strategies}

Recent advances in parallel/distributed computing, particularly cloud computing \cite{armbrust2010view,chasins2022sky,10.1145/3552309}, provide new advantages and challenges for evolutionary algorithms (EAs) \cite{eiben2015evolutionary,Schoenauer2015}. Although it comes as no surprise that parallelism is not a panacea for all cases \cite{fisher1988your,Wolpert2021,doi:10.1126/science.271.5245.56}, DES are playing an increasingly important role in large-scale BBO in the post-Moore era \cite{doi:10.1126/science.aam9744,dean2018new}. Refer to e.g., the recently proposed \textit{hardware lottery} \cite{10.1145/3467017} for insightful discussions. Note that here we focus on only model-level or application-level parallelism\footnote{Although some researchers viewed some distributed EAs as a new class of meta-heuristics, here we adopt a conservative perspective, that is, distributed EAs are seen as a performance enhancement under distributed computing.} (rather than instruction-level parallelism \cite{10.1145/3282307}).

According to \cite{rudolph2005parallel}, one of the first to parallelize ESs was \cite{bernutat1984evolution}, where an \textit{outdated} vector computer was employed in 1983. In the early days of ES research, Schwefel \cite{10.1007/978-1-4684-6389-7_46} used the classical coordinator-worker \cite{10.1007/978-3-642-95665-2_11,10.1007/3-540-61723-X_1046,10.1007/3-540-58484-6_285} model to conduct evolutionary (collective) learning of variable scalings on parallel architectures\footnote{The coordinator-worker model (aka farmer/worker \cite{10.1007/BFb0014799,eberhard2003parallel}) is typically used for computationally-intensive fitness evaluations such as optimization of aircraft side rudder and racing car rear wing \cite{10.1007/BFb0014799}), etc. The well-established Amdahl's law \cite{amdahl2007validity,amdahl2013computer} can be used as an often useful speedup estimation.}. However, only a simulated (not realistic) parallel environment was used in his experiments, where the costs of data communication, task scheduling, and distributed fault-tolerance were totally ignored. It may over-estimate the convergence performance of DES. This issue existed mainly in early DES studies such as \cite{10.1007/3-540-45712-7_41,10.1007/BFb0029754,10.1162/106365603321828970} given the fact that at that time commercial parallel/distributed computers were not widely available  and Moore's law still worked well.

Rudolph \cite{10.1007/BFb0029754} used the popular island (aka coarse-grained \cite{biscani2020parallel}) model for DES. However, it only considered the simple migration operation and did not cover the distributed self-adaptation of individual step-sizes, which often results in relatively slow convergence \cite{9762038}. Although Neumann et al. \cite{neumann2011effectiveness} provided a theoretical analysis for the migration setting, its discrete assumptions and artificially constructed functions cannot be naturally extended to continuous optimization. Overall, there have been relatively rich theoretical works (e.g., \cite{doi:10.1073/pnas.2207959120,lehre2019parallel,10.1007/978-3-642-32937-1_2,qian2019distributed}) on parallel algorithms for discrete optimization while there is little theoretical work (e.g., \cite{10.1162/evco.1993.1.4.361}) on parallel EAs for continuous optimization, up to now.

Wilson et al. \cite{10.1007/978-3-642-37192-9_52} proposed an asynchronous communication protocol to parallelize the powerful CMA-ES on cloud computing platforms. When the original CMA-ES was used as the basic computing unit for each CPU core, however, under its quadratic computational complexity the problem dimensions to be optimized are often much low (e.g., only 50 in their paper) by the modern standard for large-scale BBO. Similar issues are also found in existing libraries such as pCMALib \cite{muller2009pcmalib}, Playdoh \cite{ROSSANT2013352}, OpenFPM \cite{INCARDONA2019155} and \cite{hakkarinen2012reduced}. Glasmachers \cite{10.1145/2463372.2463424} updated strategy parameters asynchronously for Natural ES (NES) \cite{JMLR:v15:wierstra14a}, a more principal version for ES. The runtime speedup ratio obtained was below 60\% on the 8-d Rosenbrock function, which indicates the need for improvements. Reverse and Jaeger \cite{jostins2010reverse} designed a parallel island-based ES called piES to optimize a non-linear (62-d) systems biology model \cite{10.1093/bioinformatics/btm433}. In piES, only individual step-sizes were self-adapted for each island (the CMA mechanism was ignored). The speedup formulation used in their paper considered only the runtime but not the solution quality, which may lead to over-optimistic conclusions in many cases.

Recently, Cuccu et al. \cite{10.1145/3512290.3528764} proposed a DES framework called DiBB based on the partially separable assumption \cite{10.1145/3474054,duan2023cooperative}. Although it obtained significant speedups when this assumption was satisfied, DiBB did not show obvious advantages against the serial CMA-ES on ill-conditioned non-separable landscapes \cite{10.1145/3512290.3528764}.
Kucharavy et al. \cite{kucharavy2023byzantineresilient} designed a Byzantine-resilient \cite{10.1145/3335772.3335936} consensus strategy for DES. However, they did not explicitly consider the improvement of search performance and no performance data was reported in their paper. Rehbach et al. \cite{rehbach2022benchmark} used the classical (1+1)-ES as the local searcher for parallel model-based optimization on a small-sized (i.e., 16-core) parallel hardware. To our knowledge, they did not consider the more challenging distributed computing scenarios.

Another (perhaps less-known) research line for DES is Meta-ES (also referred to as Nested-ES, first proposed by Rechenberg \cite{10.1007/978-3-642-81283-5_8,rechenberg1994evolutionsstrategie}) which organizes multiple independent (parallel) ESs hierarchically \cite{10.1162/evco.2008.16.2.151,10.1145/1569901.1569970}. Although there have been relatively rich theoretical works on different landscapes (e.g., parabolic ridge \cite{10.1145/1143997.1144080}, sphere \cite{10.1145/1569901.1569971}, noisy sphere \cite{10.1145/2460239.2460242}, sharp ridge \cite{10.1145/2330163.2330208}, ellipsoid \cite{HELLWIG2016160}, conic constraining \cite{10.1145/3299904.3340306}), to our knowledge, all these theoretical models do not take overheads from distributed computing \cite{10.5555/2831090.2831104} into account. Furthermore, although these works provide valuable theoretical insights to understand Meta-ES, all the inner-ESs used in their models are relatively simple from a practical viewpoint.

In this subsection, we omit the diffusion (also called neighborhood or cellular \cite{BALUJA19931,rudolph1992parallel}) model \cite{weinert2001dynamic}, as it was rarely used in the distributed computing scenario considered in our paper. For a more general introduction to distributed EAs, refer to e.g., two recent survey papers \cite{10.1145/3377929.3389880,10.1145/3400031}.

\subsection{Large-Scale Variants of CMA-ES}

In this subsection, we review large-scale variants of CMA-ES through the lens of distributed computing. For an introduction, refer to e.g., \cite{10.1007/978-3-319-99253-2_1,10.1007/978-3-319-45823-6_1,10.1145/2908812.2908863,VARELAS2020106737} and references therein.

Because the standard CMA-ES has a quadratic time-space complexity, it is difficult to directly distribute it on cloud or clustering computing platforms for large-scale BBO. A key point to alleviate this issue is to reduce the computational complexity of the CMA mechanism, in order to fit better the (distributed) memory hierarchy. Till now, different ways have been proposed to improve its computational efficiency: 1) exploiting the low-rank structure, e.g., \cite{10.1007/978-3-319-45823-6_1,10.1145/2908812.2908863,li2017simple,li2018fast,10.1145/2001576.2001692}; 2) making the separable assumption, e.g., \cite{10.1162/evco_a_00260,10.1007/978-3-540-87700-4_30}; 3) inspiring from L-BFGS, e.g. \cite{10.1162/EVCO_a_00168,he2020mmes,loshchilov2018large}; 4) seeking computationally more efficient implementations \cite{10.1145/3377929.3389870,beyer2017simplify,NIPS2016_289dff07,10.1145/2725494.2725496,10.1145/1830483.1830556,suttorp2009efficient,10.1145/1143997.1144082}.

For many large-scale variants of CMA-ES, one obvious advantage against its standard version is their much lower time-space complexity (e.g., $O(n\log n)$ for LM-CMA). To the best of our knowledge, however, their distributed extensions are still rare up to now, despite of their clear advantages on large-scale BBO.

One key challenge for DES lies in the trade-off between (computation) simplicity and (model) flexibility. On the one hand, we need to keep the structure of CMA, which can be simply parameterized as the number of evolution paths to reconstruct, as simple as possible, in order to fit better memory hierarchy and reduce communication costs. On the another hand, we also expect to maintain the well-established invariance property as much as possible, in order to keep the flexible expressiveness/richness of model (resembling the second-order optimization method) \cite{10.1162/evco_a_00251}. To achieve such a trade-off, an efficient adaptive strategy (particularly at the meta-level) is highly desirable, which is the goal of our paper.

Our paper will enhance the state of the art from three main aspects: 1) more powerful adaptation strategies under the distributed framework as compared with existing parallel/distributed ES versions (e.g., \cite{10.1145/3512290.3528764, 10.1145/2463372.2463424, 10.1007/978-3-642-37192-9_52}); 2) efficient utilization of parallel/distributed computing resources when compared against serial large-scale ES versions (e.g., \cite{10.1145/2908812.2908863, 10.1145/2001576.2001692, 10.1145/3377929.3389870}); 3) an open-source implementation on a relatively large clustering computing platform, which is really rare in the current DES research, to our knowledge.

\section{A Multilevel Meta-Framework for DES}\label{sec:multilevel_meta_framework}

\begin{figure}
	\centering
	\includegraphics[width=0.45\textwidth,height=0.6\textwidth]{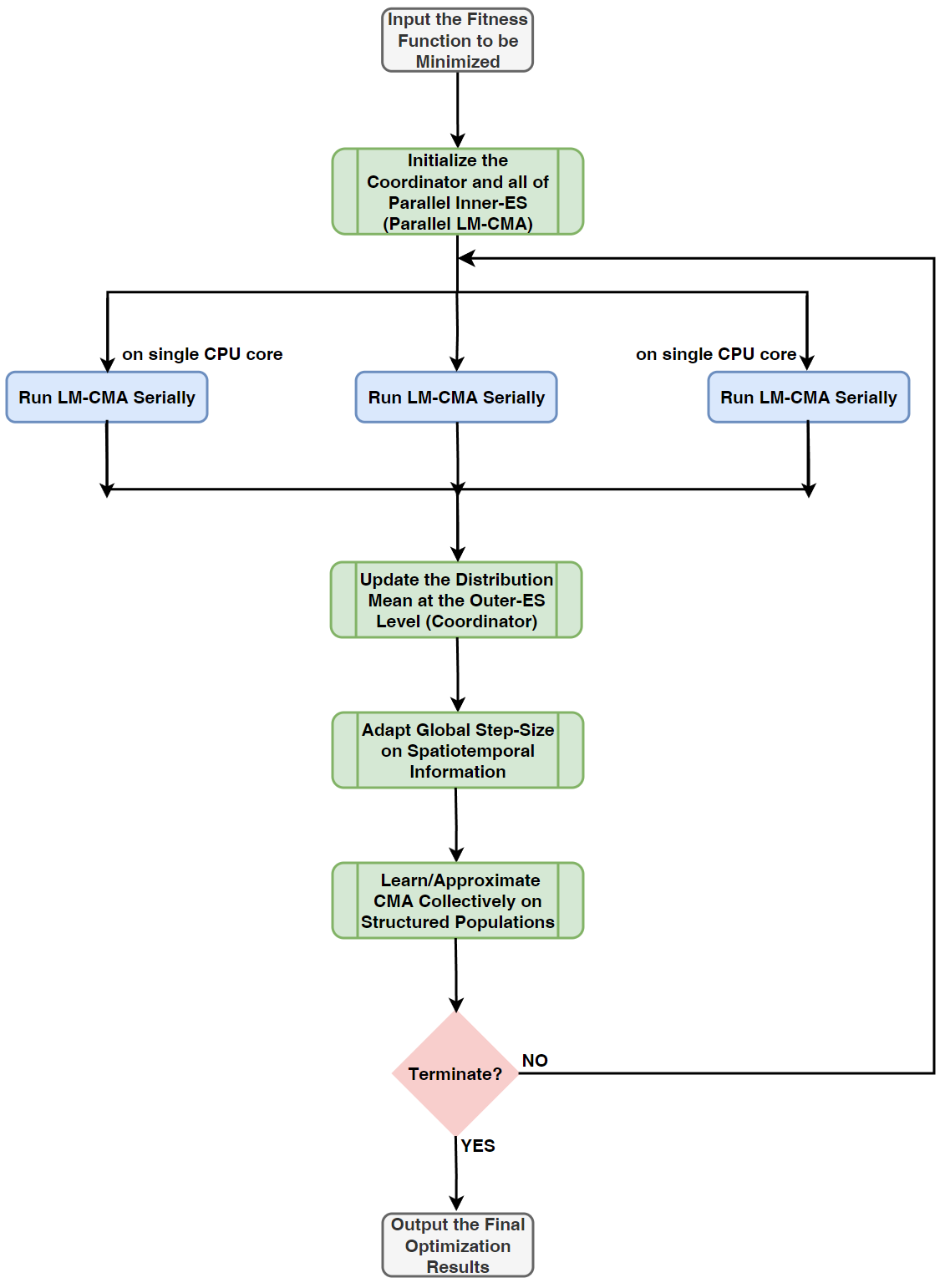}
	\caption{The flowchart diagram of our proposed approach (\hyperref[subsec:meta_framework_des]{DLMCMA}) consisting of four components: \hyperref[subsec:hierarchical_organization]{1}) hierarchical organization of LM-CMA via Meta-ES, \hyperref[subsec:update_outer_es_mean]{2}) distribution mean update at the outer-ES level, \hyperref[subsec:adapt_global_step_size]{3}) spatiotemporal global step-size adaptation, and \hyperref[subsec:collective_learning_cma]{4}) collective learning of CMA reconstruction on structured populations. }
	\label{fig:flowchart_diagram}
\end{figure}

In this paper we propose a multilevel-based meta-framework for distributed evolution strategies (DES) to minimize\footnote{Without loss of generality, the maximization problem can be easily transferred to the minimization problem by simply negating it.} the large-scale BBO problem $f(\mathbf{x}): \mathbb{R}^n \to \mathbb{R}$, where $n \gg 100$ is the dimensionality. Very recently, a research team led by the biologist Koonin has presented a mathematical model of evolution through the lens of multilevel learning in their two \textit{PNAS} papers \cite{doi:10.1073/pnas.2120037119},\cite{doi:10.1073/pnas.2120042119}. Inspired by this evolution theory, our meta-framework for DES mainly involves hierarchical organization of distributed computing units (via Meta-ES), multilevel selection\footnote{For the modern theory regarding to evolutionary transitions in individuality \cite{doi:10.1098/rsos.190202},\cite{szathmary1995major}, multilevel selection is regarded as a crucial factor to understand life's complexification \cite{okasha_2005},\cite{bozdag2023novo}.}, and collective learning of parameterized search/mutation distributions on structured populations \cite{doi:10.1126/science.1209548}. While this new evolution theory is interpreted in a mathematical way, our meta-framework, whose flowchart is shown at an relatively abstract level in Fig. \ref{fig:flowchart_diagram}, needs to be well-interpreted from an \textit{optimizer} view of point, as shown below in detail.

\subsection{Hierarchical Organization of LM-CMA via Meta-ES}\label{subsec:hierarchical_organization}

As pointed out by Rudolph, the design of DES should be well-aligned to the target hardware (see \cite{flynn1966very} for classification), in this paper we consider the clustering/cloud computing platform consisting of a number of independent high-performing Linux servers, each of which owns one shared memory and multiple CPU cores. These Linux servers are connected via a high-speed local area network (LAN).

The population structure plays a fundamental role on the search dynamics of DES \cite{alba2002parallelism}. To obtain a statistically stable learning process, we use the hierarchically organized structure from Meta-ES to control/evolve parallel LM-CMA. In principle, other LSO variants of CMA-ES could also be used here as the basic computing unit on each CPU core. When many distributed computing units are available, large populations are highly desirable for many cases such as multi-modality and noisiness \cite{PhysRevLett.130.116202,PhysRevA.107.032603,PhysRevA.107.032407}. Simply speaking, Meta-ES is an efficient way to build a structured  level for the large population, in order to reduce communication costs.

For Meta-ES, one key hyper-parameter is the isolation time $\tau$, which controls the communication frequency at different levels (i.e., between the outer-ES and all parallel inner-ESs). It is no surprise that the optimal setting of isolation time $\tau$ is problem-dependent. Generally, the longer the isolation time, the more diverse (slower) the population (local convergence); and vice versa. Furthermore, the longer the isolation time, the lower (slower) the communication cost (learning progress); and vice versa. To attain a satisfactory performance for DES, our meta-framework needs to keep reasonable trade-offs between population diversity and convergence rate, and between communication cost and learning progress, which will be tackled in the following subsections.

\subsection{Distribution Mean Update at the Outer-ES Level}\label{subsec:update_outer_es_mean}

At the outer-ES level, the \textit{elitist} \cite{10.1162/evco.2007.15.1.1,BRINGMANN201322,10.1162/evco.2010.18.1.18104} or \textit{weighted multirecombination} strategy is used to initialize the distribution mean of each inner-ES from the next isolation time, according to a controllable ratio $\mu'$ (e.g., 1/5 vs 4/5)\footnote{For simplicity, this hyper-parameter $\mu'$ is also used for the weighted multirecombination strategy of the outer-ES. See (1) for details.}. The rationale behind this parallel update strategy is presented in the following: If only the \textit{elitist} strategy is used, the parallel search process may suffer from stagnation; if only the \textit{multirecombination} strategy is used, the parallel search process may suffer from the
regression issue on some functions (e.g., with a predominated search direction). Note that for simplicity, the default distribution mean update is used for each inner-ES as the same as LM-CMA within each isolation time $\tau$.

In the outer-ES, the \textit{weighted multirecombination} update of its distribution mean $m'$ after each isolation time is mathematically calculated as
\begin{align}
	\mathbf{m}'=\sum_{i=1}^{\mu'}w_{i;\lambda'}'\mathbf{m}_{i;\lambda'}, where \sum_{i=1}^{\mu'}w_{i;\lambda'}'=1,
\end{align}
where $\mu'$ is the used (selected) number of all ($\lambda'$) parallel inner-ESs, $w_{i;\lambda'}'$ and $\mathbf{m}_{i;\lambda'}$ are the weight and distribution mean of the $i$th-ranked\footnote{For minimization, the lower the fitness (cost), the higher the ranking.} inner-ES, respectively. Even at the outer-ES level, we still follow the standard practice (that is, the higher the ranking, the larger the weight) to set all the weights. Refer to e.g., Arnold's theoretical analysis \cite{ARNOLD200618} for a better understanding.

\subsection{Spatiotemporal Global Step-Size Adaptation (STA)}\label{subsec:adapt_global_step_size}

As previously pointed out by Rudolph \cite{rudolph1992correlated}, the (online) control problem of strategy parameters is multi-modal and noisy. Even if the optimization problem is deterministic rather than noisy, the random sampling nature from the inner-ES level makes it difficult to adapt the global step-size at the outer-ES level. To obtain a reliable estimation, a \textit{relatively long} isolation time $\tau$ may be preferred, which also brings some extra benefits w.r.t. communication costs and fault tolerance.

In order to exploit both the \textit{spatio} and \textit{temporal} (non-local) information on-the-fly, our meta-framework combines the less-known Meta-ES strategy with the well-known CSA strategy (or its recent population-based or success-based variants for LSO)\footnote{In this paper, we do not modify the CSA-style strategy for all inner-ESs. Instead, we focus on the global step-size adaptation at the outer-ES level, which is crucial for DES.}. Specifically, three parallel design strategies are used at the outer-ES to balance adaptation speed and meta-population diversity\cite{10.1007/BFb0029753}: 1) when the elitist strategy is used for the distribution mean update, the same \textit{elitist} strategy is also used to update the global step-size of the corresponding inner-ES to obtain the relatively stable evolution process; 2) given a predefined proportion (e.g., 1/5), some inner-ESs \textit{mutate} the weighted multi-recombined global step-size $\sigma'$ as $\sigma_i\sim \mathcal{U}(\sigma'*a, \sigma'*b)$, where $0<a<1<b$, implicitly based on the \textit{strong causality} assumption \footnote{For simplicity, in this paper we follow the common suggestion from Meta-ES and set $a=0.3$ and $b=3.3$, respectively (note that $1/3.3\approx0.3$ leads to unbiasedness in the logarithmic scale).}; 3) otherwise, the global step-size will be \textit{uniformly} sampled from a reasonable search range\footnote{In practice, a reasonable search range seems to be easier to set than a reasonable value.}, in order to maintain diversity and reduce the risk of getting trapped into local minima.

The weighted multirecombination for the global step-size $\sigma'$ of the outer-ES should be done (somewhat) in an \textit{unbiased} way:
\begin{align}
	\sigma'=\sum_{i=1}^{\mu'}\frac{w_{i;\lambda'}'\sigma_{i;\lambda'}}{\sqrt{\sum_{i=1}^{u'}w_{i;\lambda'}'}},
\end{align}
where the denominator $\sqrt{\sum_{i=1}^{u'}w_{i;\lambda'}'}$ ensures $\sigma' \sim \mathcal{N}(0,1)$ at the logarithmic scale under neutral selection (one of basic design principles from the ES community). Different from the CSA, STA does not use the exponential smoothing method at the outer-ES level, since the temporal information has been well exploited by each inner-ES and it is hard to set the corresponding learning rate and decaying factor (undoubtedly, it is expensive to set them at the outer-ES level).

\subsection{Collective Learning of CMA on Structured Populations}\label{subsec:collective_learning_cma}

The most prominent feature of CMA-ES appears to be its adaptive encoding (i.e., invariance against affine transformation) ability, especially on non-separable, ill-conditioned problems. As a general-purpose black-box optimizer, we expect DES to keep this powerful feature as much as possible. In order to be communication-efficient, however, we need to properly compress the standard $n \times n$ covariance matrix to fit the distributed shared memory; but this may destroy the highly desirable invariance property.
In this paper, we choose to use one of its large-scale variants (i.e., LM-CMA) as the basic computing unit on each CPU core, in order to reduce the communication cost after each isolation time $\tau$.

The simplified form of CMA, derived by Beyer and Sendhoff \cite{beyer2017simplify}, is presented as
\begin{equation}
	\begin{aligned}
		\mathbf{C}^{t+1} \leftarrow & \mathbf{C}^t \Big\{\mathbf{I} + \frac{c_1}{2}(\mathbf{p}^{t+1}(\mathbf{p}^{t+1})^T - \mathbf{I})                    \\
		                            & + \frac{c_\mu}{2}(\sum_{i=1}^{\mu}{w_i \mathbf{z}_{i;\lambda}^t{(\mathbf{z}_{i;\lambda}^t)}^T} - \mathbf{I})\Big\},
	\end{aligned}
\end{equation}
where $\mathbf{C}^t$ is the transformation matrix at the $t$-th generation (another form of the covariance matrix to avoid eigen-decomposition), $\mathbf{I}$ is the identity matrix, $\mathbf{p}^{t+1}$ is the evolution path at the $(t+1)$-th iteration, $w_i$ is the weight for the $i$-th ranked individual, $\mathbf{z}_{i;\lambda}^{t}$ is the realized random sample from the standard normal distribution for the $i$-th ranked individual, $\mu$ is the number of parents of the inner-ES, $c_{1}$ is the coefficient of the rank-one update \cite{10.1162/106365601750190398}, and $c_{\mu}$ is the coefficient of the rank-$\mu$ update \cite{10.1162/106365603321828970}, respectively.

After omitting the update-$\mu$ update, the sampling procedure can be significantly reduced to

\begin{equation}
	\begin{aligned}
		d_i^t= & \Big((1-\frac{c_1}{2})\mathbf{I}+\frac{c_1}{2}\mathbf{p}^1{(\mathbf{p}^1)}^T\Big)                      \\
		       & \times\Big((1-\frac{c_1}{2})\mathbf{I}+\frac{c_1}{2}\mathbf{p}^2{(\mathbf{p}^2)}^T\Big)                \\
		       & \times \cdots                                                                                          \\
		       & \times\Big((1-\frac{c_1}{2})\mathbf{I}+\frac{c_1}{2}\mathbf{p}^{t-1}(\mathbf{p}^{t-1})^T\Big)          \\
		       & \times\Big((1-\frac{c_1}{2})\mathbf{I}+\frac{c_1}{2}\mathbf{p}^t{(\mathbf{p}^t)}^T\Big)\mathbf{z}_i^t.
	\end{aligned}
\end{equation}

It is worthwhile noting that the above equation should be calculated from right to left, in order to get a linear complexity for each operation. Owing to the limit of pages, please refer to \cite{loshchilov2018large} for detailed mathematical derivations. To reduce the overall computational complexity, only a small amount of evolution paths (parameterized as $n^{e}$ here) are used in all limited-memory LSO variants but with different selection rules. Because the successive evolution paths usually exhibit relatively high correlations, a key point is to make a diverse baseline of evolution paths for the covariance matrix reconstruction. In practice, different problems often have different topology structures and need different fitting structures, which naturally lead to the structure learning problem.

For properly approximating CMA under distributed computing, our meta-framework employs two adaptive distributed strategies for structure and distribution learning, respectively. First, as the structure learning often operates at a relatively slow-changing scale and controls the richness of distribution model, we implicitly adapt the total number of reconstructed evolution paths $n^{e}$ via the \textit{elitist} strategy, in order to obtain a reliable learning progress at the outer-ES level. In other words, we save a certain  elitist ratio\footnote{It is set to $\mu'$ for consistency and simplicity in this paper.} as some (but not all) parallel inner-ESs for the next isolation time and for the remaining parallel inner-ESs we sample  $n^{e}$ \textit{uniformly} in a reasonable setting range\footnote{In practice, the setting of the search range of $n^{e}$ depends on the available memory in the used distributed computing platform, which is easy to obtain.}, in order to maintain the diversity of structure learning at the outer-ES level. Note that for each inner-ES, its reconstructed structure is always fixed within each isolation time.

For collective learning of search distributions under distributed structured populations, our meta-framework uses a simple yet efficient \textit{weighted multirecombination} strategy to combine the evolution paths from the elitist inner-ESs into a shared pool of reconstructed evolution paths $\mathbf{P}'$ for the next isolation time. Cautiously, owing to possibly heterogeneous shapes from structure learning, we need to align the weighted multirecombination operation as follows (first $\mathbf{p}'$ is initialized as a $\underset{i=1,...,\mu'}{\max}(n^{e}_i) \times n$ zero matrix after each isolation time):

\begin{align}
	\mathbf{P}'[-n^{e}_i:]\mathrel{{+}{=}}\frac{w_{i;\lambda'}'}{\sqrt{\sum_{i=1}^{u'}w_{i;\lambda'}'}}\mathbf{P}_i[-n^{e}_i:] (i=1,\cdots,\mu'),
\end{align}

\noindent where $[-n^{e}_i:]$ denotes all indexes starting from the last $n^{e}_i$ column to the end, $\sqrt{\sum_{i=1}^{u'}w_{i;\lambda'}'}$ ensures unbiasedness under neutral selection, and $\mathbf{p}_i$ is a pool of reconstructed evolution paths from the $i$-th ranked inner-ES, respectively.

\begin{algorithm}[!t]
	\caption{A Multilevel-based Meta-Framework for DES.}\label{alg:alg1}
	\begin{algorithmic}[1]
		\Require{\(\lambda'\): the number of all parallel inner-ESs (LM-CMAs)}
		\Statex{\(\mu'\): the number of elitists for the outer-ES}
		\Statex{\(\mathbf{m}_i\): the distribution mean of the \(i\)-th inner-ES}
		\Statex{\(\sigma_i\): the global step-size of the \(i\)-th inner-ES}
		\Statex{\(\mathbf{P}_i\): a pool of evolution paths of the \(i\)-th inner-ES}
		\Statex{\(\tau \): the isolation time (i.e., runtime of each LM-CMA)}
		\Statex{\(\sigma'\): the global step-size of the outer-ES}
		\Statex{\(\sigma_{\max}\): maximally possible value of global step-size}
		\Statex{\(n_i^e\): number of evolution paths for \(i\)-th inner-ES}

		\Ensure{\(\mathbf{x}^*\): the best-so-far solution}
		\Statex{\(f^*\): the best-so-far fitness (cost)}

		\While{the \textit{maximal} runtime is not reached}

		\LComment{do a \textcolor{red}{\textit{parallel for}} loop over inner-ESs (LM-CMAs)}
		\For{\(i=1\) to \(\lambda'\)}
		\If{\(i \mathrel{<=}  \mu'\)} \Comment{use elitist for part inner-ESs}
		\State{\(\mathbf{m}_i,\sigma_i,\mathbf{P}_i,\mathbf{x}^*_i, f^*_i \gets \textbf{LM-CMA}(\mathbf{m}_{i;\lambda},\sigma_{i;\lambda},\mathbf{P}_{i;\lambda}, \tau)\)}
		\Else \Comment{on the multi-recombination strategy}
		\If{\(i \mathrel{<=} \mu' + (\lambda' - \mu')/5\)} \Comment{for Meta-ES}
		\State{\(\sigma \gets U(0.3\sigma', 3.3\sigma')\)} \Comment{mutate step-size}
		\Else \Comment{for step-size diversity}
		\State{\(\sigma \gets U(0, \sigma_{\max})\)} \Comment{uniformly sample}
		\EndIf

		\State{\(n_i^e \gets U(n_{\min}^e, n_{\max}^e)\)} \Comment{uniformly sample}
		\State{\(\mathbf{m}_i,\sigma_i,\mathbf{P}_i,\mathbf{x}^*_i, f^*_i \gets \textbf{LM-CMA}(\mathbf{m}', \sigma, \mathbf{P}', \tau, n_{i}^e)\)}
		\EndIf
		\EndFor

		\State{\(\mathbf{m}' \gets \sum_{i=1}^{\mu'}w_{i;\lambda'}'\mathbf{m}_{i;\lambda'}\)} \Comment{update distribution mean}
		\State{\(\sigma' \gets \sum_{i=1}^{\mu'}\frac{w_{i;\lambda'}'\sigma_{i;\lambda'}}{\sqrt{\sum_{i=1}^{u'}w_{i;\lambda'}'}}\)} \Comment{multi-recombine for STA}

		\LComment{collective learning on distributed populations}
		\State{\(\mathbf{P}' \gets \mathbf{0}_{[\underset{i=1,...,\mu'}{\max}(n^{e}_i) \times n]}\)} \Comment{\(\max \) for shape alignment}
		\For{\(i=1, \dots, \mu'\)} \Comment{only consider elitist}
		\State{\(\mathbf{P}'[-n^{e}_i:]\mathrel{{+}{=}}\frac{w_{i;\lambda'}'}{\sqrt{\sum_{i=1}^{u'}w_{i;\lambda'}'}}\mathbf{P}_i[-n^{e}_i:]\)}
		\EndFor

		\State{\(\mathbf{x}^* \gets \min(\mathbf{x}^*, \mathbf{x}^*_{1},...,\mathbf{x}^*_{\lambda'})\)} \Comment{update the best solution}
		\State{\(f^* \gets \min(f^*, f^*_{1},...,f^*_{\lambda'})\)} \Comment{update the best fitness}

		\EndWhile
	\end{algorithmic}
\end{algorithm}

\subsection{A Meta-Framework for DES}\label{subsec:meta_framework_des}

Here we combine all the aforementioned design choices into our distributed meta-framework, as presented in \textbf{Algorithm 1}\footnote{Rinnooy Kan and Timmer \cite{rinnooy1987stochastic} from the mathematical programming community considered a multi-level method for stochastic optimization in the context of \textit{single linkage}. However, our approach is orthogonal to their method.}. To implement our meta-framework, we select one state-of-the-art clustering computing software called \textit{ray}\cite{10.5555/3291168.3291210} as the key engine of distributed computing\footnote{\tiny \href{https://github.com/Evolutionary-Intelligence/M-DES/blob/main/README.md\#troubleshooting-tips}{https://github.com/Evolutionary-Intelligence/M-DES/blob/main/README.md\#troubleshooting-tips}}. As compared with other existing distributed computing systems such as MPI \cite{10.1007/3-540-45356-3_40}, P2P \cite{10.1007/3-540-45712-7_64,biazzini2010gossiping}, MapReduce \cite{10.1145/1327452.1327492,10.1162/evco_a_00213}, Spark \cite{JMLR:v17:15-237}, BlockChain \cite{bizzaro2020proof}, \textit{ray} provides a flexible programming interface for Python and a powerful distributed scheduling strategy to cater to modern challenging AI applications such as population-based training \cite{pbt-icml,jaderberg2017population}, AutoML \cite{pmlr-v119-real20a}, and open-ended learning/evolution \cite{pmlr-v119-wang20l}. Owing to the intrinsic complexity of distributed algorithms, we provide an open-source Python implementation for our proposed meta-framework available at \url{https://github.com/Evolutionary-Intelligence/M-DES}, in order to ensure repeatability and benchmarking\cite{10.1145/3466624}.

For simplicity and ease to analyze, our meta-framework uses the \textit{generational} (rather than \textit{steady-state}) population update strategy at the outer-ES level. Although typically the steady-state method could maximize the parallelism level especially for heterogeneous environments, the asynchronous manner makes distributed black-box optimizers difficult to debug. In this paper, we consider only the generational population update manner, since it makes the updates and communications of search/mutation distributions easier to understand and analyze under distributing computing.

\section{Large-Scale Numerical Experiments}\label{sec:exp}

To study the benefits (and costs) of our proposed meta-framework (simply named as DLMCMA here), we conduct numerical experiments on a set of large-scale benchmarking functions with memory-expensive fitness evaluations, arguably reflecting many challenging data-driven optimization problems. To ensure repeatability\footnote{\tiny \href{https://github.com/Evolutionary-Intelligence/M-DES/blob/main/README.md\#step-by-step-instructions}{https://github.com/Evolutionary-Intelligence/M-DES/blob/main/README.md\#step-by-step-instructions}} and promote benchmarking \cite{WHITLEY1996245}, a set of involved experimental data and Python code are openly available at our companion website \url{https://github.com/Evolutionary-Intelligence/M-DES}.

\subsection{Experimental Settings}

\textbf{Test Functions:} We choose a set of 13 commonly used test functions, as shown in Table \ref{tab:benchmark_functions} for their mathematical formula (see e.g., COCO/BBOB \cite{10.1145/1830761.1830790} or NeverGrad \cite{10.1145/3460310.3460312} for implementations). These functions can be roughly classified to two families (i.e., unimodal and multimodal functions) to compare \textit{local} and \textit{global} search abilities, respectively. For benchmarking large-scale BBO, the dimensions of all the test functions are set to 2000. We also use the standard angle-preserving (i.e., rotation) transformation \cite{10.1162/106365601750190398} (rather than \cite{VARELAS2020106737}) and random shift to generate non-separability and avoid the origin as the global optimum, respectively. This involved matrix-vector multiplication operator results in the memory-expensive fitness evaluation, arguably one significant feature of many real-world data-driven optimization problems. In order to speedup parallel fitness evaluations, we use the simple yet efficient \textit{shared memory} trick for our distributed optimizer.

\begin{table}[!t]
	\centering
	\caption{A Set of 13 Benchmarking Functions}\label{tab:benchmark_functions}
	\resizebox{0.485\textwidth}{!}{%
		\begin{tabular}{c|l|l}
			\toprule[1.5pt]
			 & \makecell[c]{\textbf{Name}} & \makecell[c]{\textbf{Expression}}                                                                                \\
			\midrule[1pt]
			\multirowcell{8}{\textbf{unimodal}                                                                                                                \\ \textbf{(local)}} & {Sphere}          & $ f(x) = \sum_{i = 1}^{n} x_i^2$ \\
			 & {Cigar}                     & $ f(x) = x_1^2 + 10^6 \sum_{i = 2}^{n} x_i^2 $                                                                 \\
			 & {Discus}                    & $ f(x) = 10^6 x_1^2 + \sum_{i = 2}^{n} x_i^2 $                                                                   \\
			 & {Ellipsoid}                 & $ f(x) = \sum_{i = 1}^{n} 10^{\frac{6(i- 1)}{n - 1}} x_i^2 $                                                     \\
			 & {DifferentPowers}           & $ f(x) = \sum_{i = 1}^{ n} \left | x_i \right | ^{\frac{2 + 4(i - 1)}{n - 1}} $                                  \\
			 & {Schwefel221}               & $ f(x) = \max(\left | x_1 \right |, \cdots, \left | x_n \right |) $                                              \\
			 & {Step}                      & $ f(x) = \sum_{i = 1}^{n}(\lfloor x_i + 0.5 \rfloor)^2 $                                                           \\
			 & {Schwefel12}                & $ f(x) = \sum_{i=1}^{n}(\sum_{j=1}^{i}x_j)^2 $                                                                   \\
			\midrule[1pt]
			\multirowcell{5}{\textbf{multimodal}                                                                                                              \\ \textbf{(global)}} & {Ackley}          & $ f(x) = -20 e^{-0.2 \sqrt{\frac{1}{n} \sum_{i = 1}^{n} x_i^2}} - e^{\frac{1}{n} \sum_{i = 1}^{n} \cos(2 \pi x_i)} + 20 + e $ \\
			 & {Rastrigin}                 & $ f(x) = 10 n + \sum_{i = 1}^{n} (x_i^2 - 10 \cos(2 \pi x_i)) $                                                  \\
			 & {Michalewicz}               & $ f(x) = -\sum_{i=1}^{n}\sin(x_i)(\sin(\frac{ix_i^2}{\pi})){}^{20} + 600$                                        \\
			 & {Salomon}                   & $ f(x) = 1 - \cos(2\pi\sqrt{\sum_{i=1}^{n}x_i^2}) + 0.1 \sqrt{\sum_{i=1}^{n}x_i^2}$                              \\
			 & {ScaledRastrigin}           & $ f(x) = 10 n + \sum_{i = 1}^{n} ((10^{\frac{i - 1}{n - 1}} x_i)^2 -10\cos(2\pi 10^{\frac{i - 1}{n - 1}} x_i)) $ \\
			\bottomrule[1.5pt]
		\end{tabular}%
	}
\end{table}%

\begin{figure*}
	\centering
	\includegraphics[width=0.95\textwidth, height=0.7\textwidth]{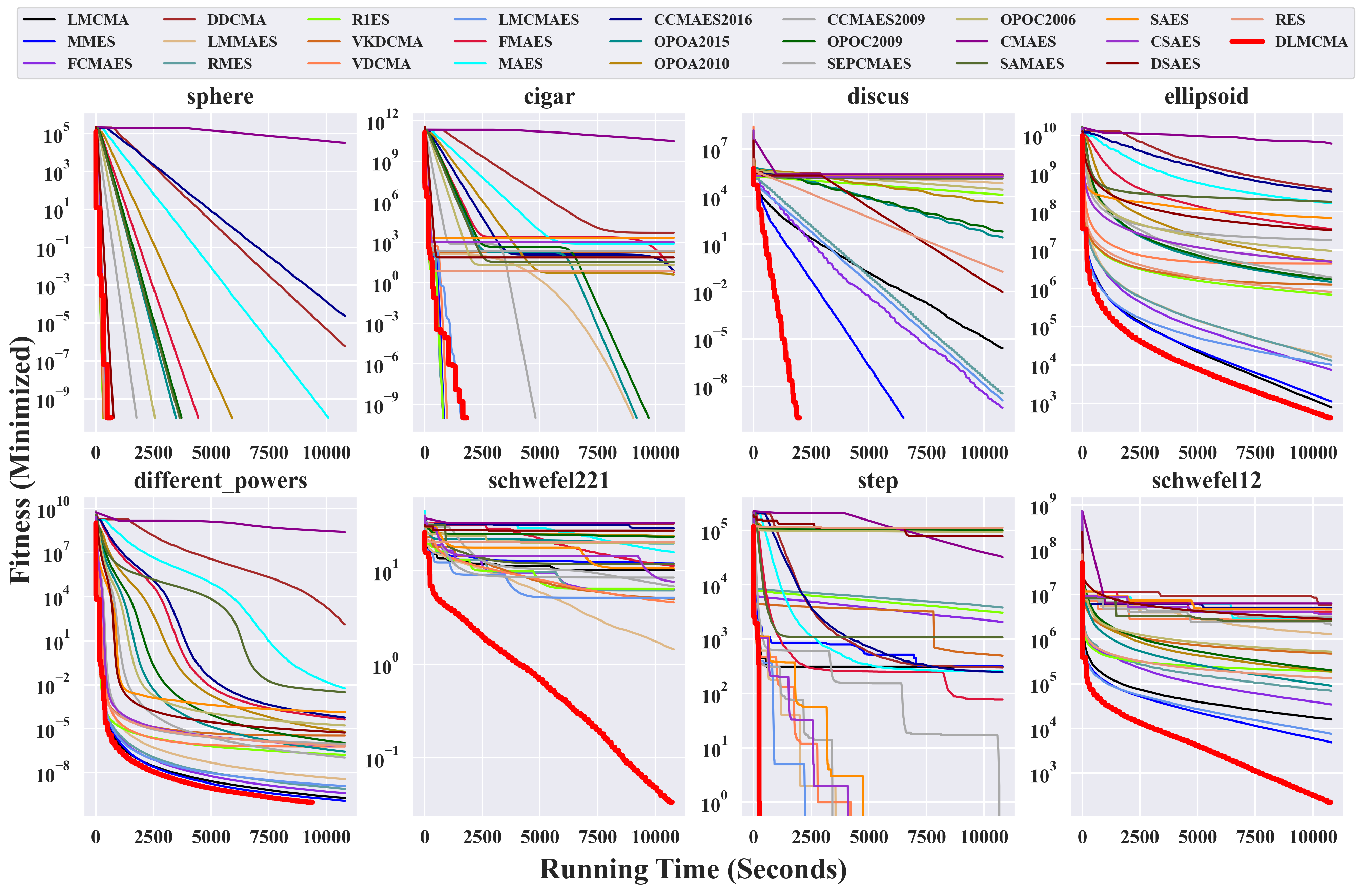}
	\caption{Median convergence curves on a set of 2000-d \textit{unimodal} functions given the maximal runtime (3 hours) and the cost threshold ($1e^{-10}$).}\label{fig:median_unimodal_1}
\end{figure*}

\begin{figure*}
	\centering
	\includegraphics[width=0.95\textwidth, height=0.7\textwidth]{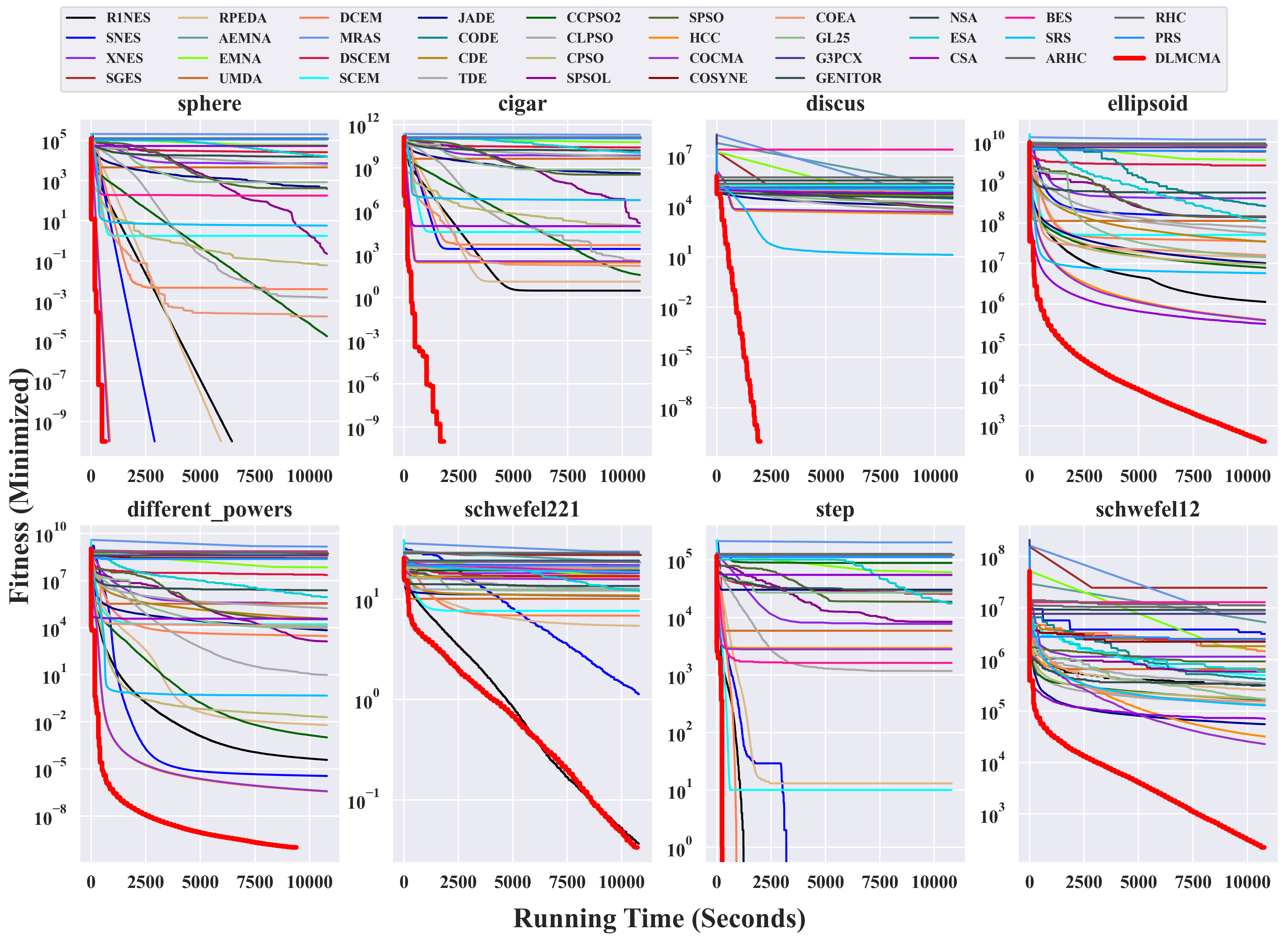}
	\caption{Median convergence curves on a set of 2000-d \textit{unimodal} functions given the maximal runtime (3 hours) and the cost threshold ($1e^{-10}$).}\label{fig:median_unimodal_2}
\end{figure*}

\textbf{Benchmarking Optimizers:} To benchmark the advantages and disadvantages of different approaches, we select a total of 61 black-box optimizers from different families (i.e., Evolution Strategies - ES, Natural Evolution Strategies - NES, Estimation of Distribution Algorithms - EDA, Cross-Entropy Method - CEM, Differential Evolution - DE, Particle Swarm Optimizer - PSO, Cooperative Coevolution - CC, Simulated Annealing - SA, Genetic Algorithms - GA, Evolutionary Programming - EP, Pattern Search - PS, and Random Search - RS) implemented in a recently well-designed Python library called \href{https://github.com/Evolutionary-Intelligence/pypop}{\textbf{PyPop7}}. Due to the page limit, here we only use their \textit{abbreviated} (rather than \textit{full}) names to avoid legend confusion when plotting the convergence figures. For implementation details about all these optimizers and their hyper-parameter settings, please refer to this online website \url{pypop.rtfd.io} and references therein. Overall, these optimizers are grouped into two classes for plotting convergence curves: ES-based optimizers and others, as shown in Fig. (\ref{fig:median_unimodal_1}, \ref{fig:median_multimodal_1}) and Fig. (\ref{fig:median_unimodal_2}, \ref{fig:median_multimodal_2}), respectively.

\textbf{Computing Environments:} We set the parallel number of inner-ESs ($\lambda'$) to 380 and 540 (CPU cores) for our DLMCMA on unimodal and multimodal functions, respectively. On multimodal functions, more parallel inner-ESs often bring larger population diversity. Although the optimal setting of $\lambda'$ is problem-dependent, we do not fine-tune it for each function. We empirically set the isolation time $\tau$ to 150 seconds on all functions though this is \textit{not necessarily} an optimal value for each function.

Owing to the time-consuming experiment process for LSO, We run each optimizer 10 and 4 times on each unimodal and multimodal function, respectively. The total CPU single-core runtime needed in our experiments is estimated up to \textbf{18600 hours}, that is, 775 days = $(10 \times 8 \times 3 + 4 \times 5 \times 3) \times 62$ hours.

\subsection{Comparing Local Search Abilities}

On the \textit{sphere} function, arguably one of the simplest test cases for continuous optimization, it is highly expected that the optimizer could obtain a fast rate of convergence. Most ES-based optimizers could obtain a satisfactory (not necessarily optimal) performance except some CMA-ES variants with quadratic complexity (Fig.~\ref{fig:median_unimodal_1}). For quadratic-complexity CMA-ES variants (e.g., DD-CMA \cite{10.1162/evco_a_00260}, MA-ES \cite{10.1145/3377929.3389870}, and C-CMA-ES \cite{NIPS2016_289dff07}), the overall runtime is dominated heavily by the CMA mechanism rather than the function evaluation time, therefore resulting in a much lower adaptation speed.

There is one predominated search direction needed to be explored for the \textit{cigar} function. This means that a low-rank learning strategy (e.g., R1-ES \cite{li2017simple}) is typically enough to capture the main direction via adaptation. Owing to the extra cost brought from distributed computing, our meta-framework (DLMCMA) obtains a slightly slow convergence speed as shown in Fig.~\ref{fig:median_unimodal_1} (reflecting limitations of parallelism \cite{markov2014limits}). However, it could approximate the low-rank learning ability well, given that the initial number of reconstructed evolution paths does not match the optimal setting.

For both functions \textit{discus} and \textit{ellipsoid}, there exist multiple promising search directions (see their relatively even eigenvalue distributions). Therefore, a much richer reconstruction model is preferred for CMA. On the \textit{discus} function (Fig.~\ref{fig:median_unimodal_1}), our DLMCMA could show the $>$3x runtime speedup w.r.t. the second ranked optimizer (i.e., MM-ES \cite{he2020mmes}). On the \textit{ellipsoid} function, our DLMCMA nearly always shows the best convergence speed during evolution (Fig.~\ref{fig:median_unimodal_1}), because its collective learning strategy maintains the better diversity of reconstructed evolution paths via utilizing the distributed computing resource. Interestingly, similar observations could also be found in another two challenging functions (i.e., \textit{differentpowers} and \textit{schwefel12}) with multiple search directions (Fig.~\ref{fig:median_unimodal_1}).

For both \textit{schwefel221} and \textit{step} functions, there are a large number of plateaus in high-dimensional cases, which result in a rugged fitness landscape. For ESs, a key challenge is to properly adapt the global step-size to pass these plateaus\footnote{In other words, this needs to find an appropriate \textit{evolution window}, popularized by Rechenberg (one of the evolutionary computation pioneers).}. Luckily, our meta-framework can tackle this challenge well and can achieve the best convergence rate on both of them (Fig.~\ref{fig:median_unimodal_1}), since our STA strategy could keep the diversity of the global step-size well while avoiding to diverge it (with the help of the elitist strategy)\footnote{The punctuated-equilibria-style convergence is dated back to at least \cite{cohoon2013punctuated}, depending on the used viewpoint (as pointed out by Schwefel, one of the evolutionary computation pioneers).}.

As we can clearly see from Fig.~\ref{fig:median_unimodal_2}, our DLMCMA could always obtain the best convergence rate when compared with all other algorithm families except that on \textit{schwefel221} R1-NES shows a very similar performance. Overall, our DLMCMA achieves the best or competitive performance on these unimodal functions, validating the benefits of multilevel distributed learning empirically.

\subsection{Comparing Global Search Abilities}

\begin{figure}
	\centering
	\includegraphics[width=0.48\textwidth, height=0.9\textwidth]{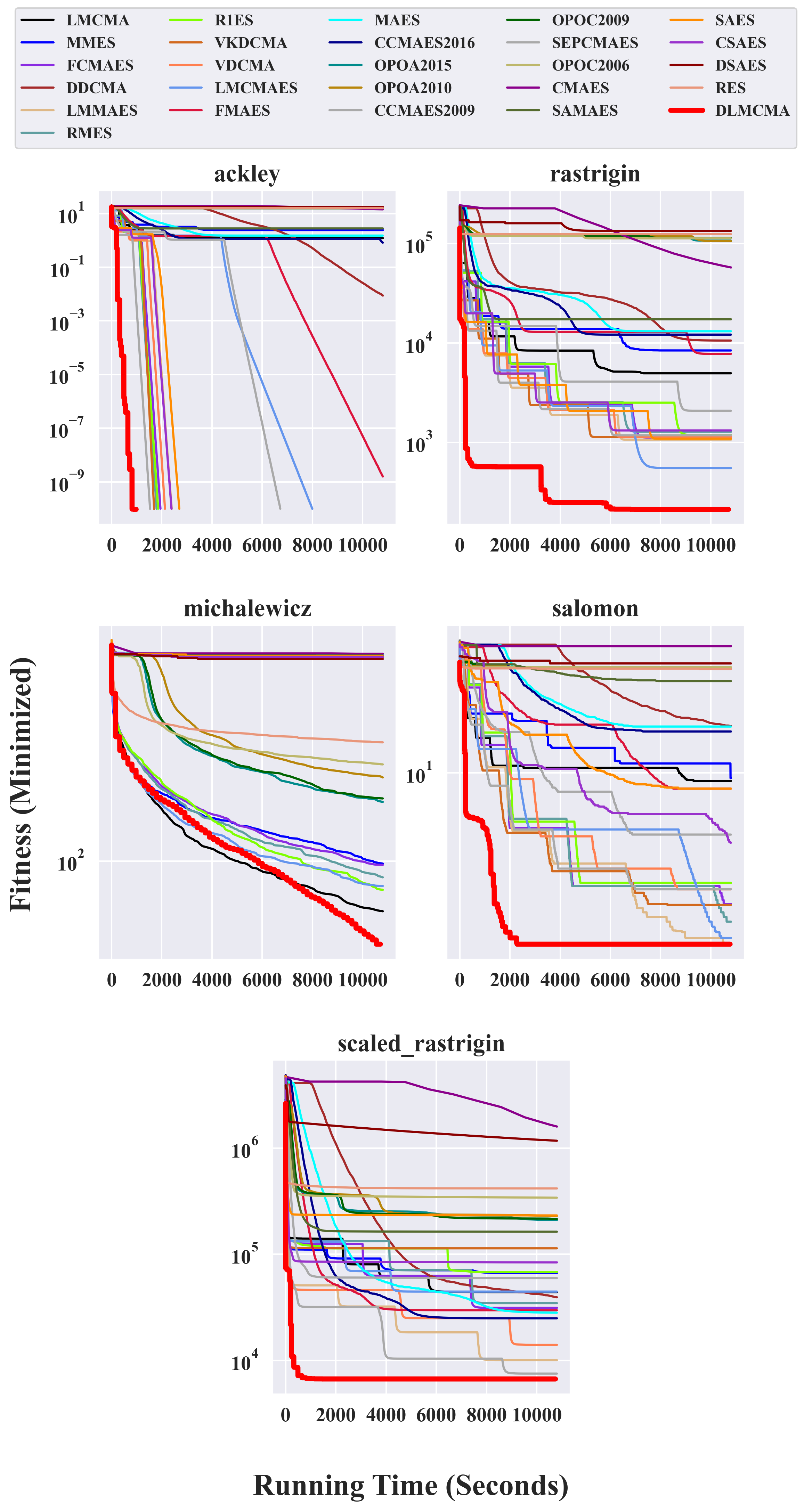}
	\caption{Median convergence curves on a set of 2000-d \textit{multimodal} functions given the maximal runtime (3 hours) and the cost threshold ($1e^{-10}$).}\label{fig:median_multimodal_1}
\end{figure}

\begin{figure}
	\centering
	\includegraphics[width=0.49\textwidth, height=0.9\textwidth]{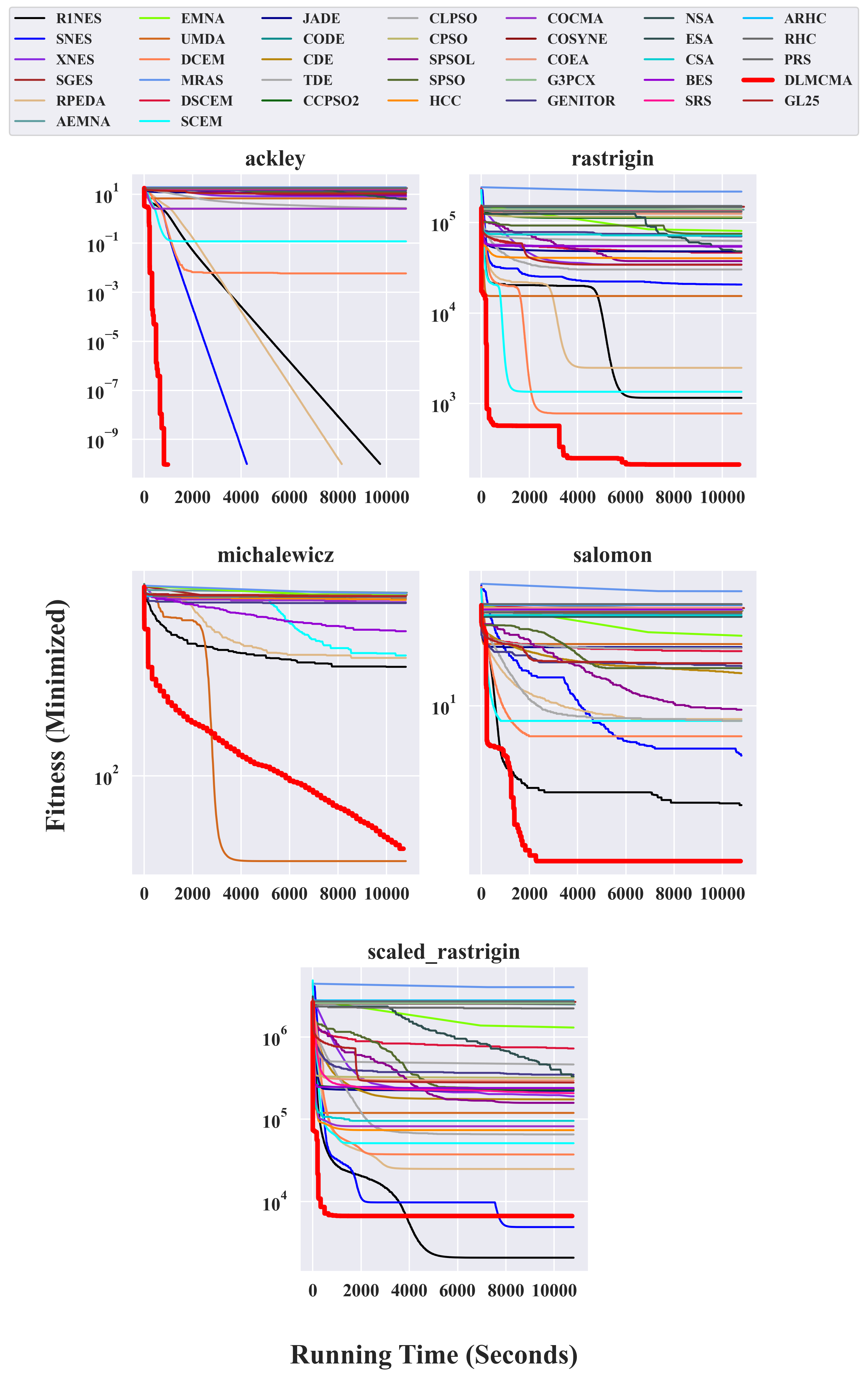}
	\caption{Median convergence curves on a set of 2000-d \textit{multimodal} functions given the maximal runtime (3 hours) and the cost threshold ($1e^{-10}$).}\label{fig:median_multimodal_2}
\end{figure}

For minimizing the \textit{ackley} function, it seems to be looking for ``a needle in the haystack". However, there is a global landscape structure to be available, which can be used to accelerate the global convergence rate of the optimizer (if well-utilized). We find that many large-scale variants of CMA-ES utilize this property, even under a small population setting. Our DLMCMA can also approximate this global structure well, therefore achieving the best convergence rate after bypassing all (shallow) local optima (Fig.~\ref{fig:median_multimodal_1} and \ref{fig:median_multimodal_2}).

On the classic \textit{rastrigin} function, there exist a large number of relatively deep local minima, which can hinder the optimization process. To escape from these local optima, the simple yet efficient restart \cite{10.1007/978-3-540-30217-9_29} strategy from CMA-ES will increase the number of offspring after each restart. Clearly, our DLMCMA obtains much better results among all optimizers, with the help of multiple restarts (Fig.~\ref{fig:median_multimodal_1} and \ref{fig:median_multimodal_2}).

Our proposed DLMCMA obtains the second ranking only after UMDA \cite{larranaga2001estimation} on the \textit{michalewicz} function (Fig.~\ref{fig:median_multimodal_1} and \ref{fig:median_multimodal_2}), which seems to have a relatively weak global structure. The default population size of UMDA is relatively large (200) while that of each (local) LM-CMA used in our DLMCMA is small by default. Despite this difference, our DLMCMA still can drive the parallel evolution process over structured populations to approach the best after 3 hours.

On the multimodal function \textit{salomon}, our DLMCMA finds the best solution much faster than all other optimizers (Fig.~\ref{fig:median_multimodal_1} and \ref{fig:median_multimodal_2}). However, the restart strategy cannot help to find a better solution, which may indicate that this found solution is near a deep local optimum. On another multimodal function \textit{scaledrastrigin}, our DLMCMA ranks the third, only after R1-NES and SNES \cite{JMLR:v15:wierstra14a} (Fig.~\ref{fig:median_multimodal_1} and \ref{fig:median_multimodal_2}). We notice that the parallel evolution process stagnates even at the early stage, which means that we need a better restart strategy for this function. We leave it for future work.

In summary, our meta-framework achieves the very competitive performance on both unimodal and multimodal test functions considered in the experiments, under the challenging distributed computing scenarios.

\subsection{Overhead Analysis of Memory Communications}

\begin{figure}
	\centering
	\includegraphics[width=0.48\textwidth, height=0.48\textwidth]{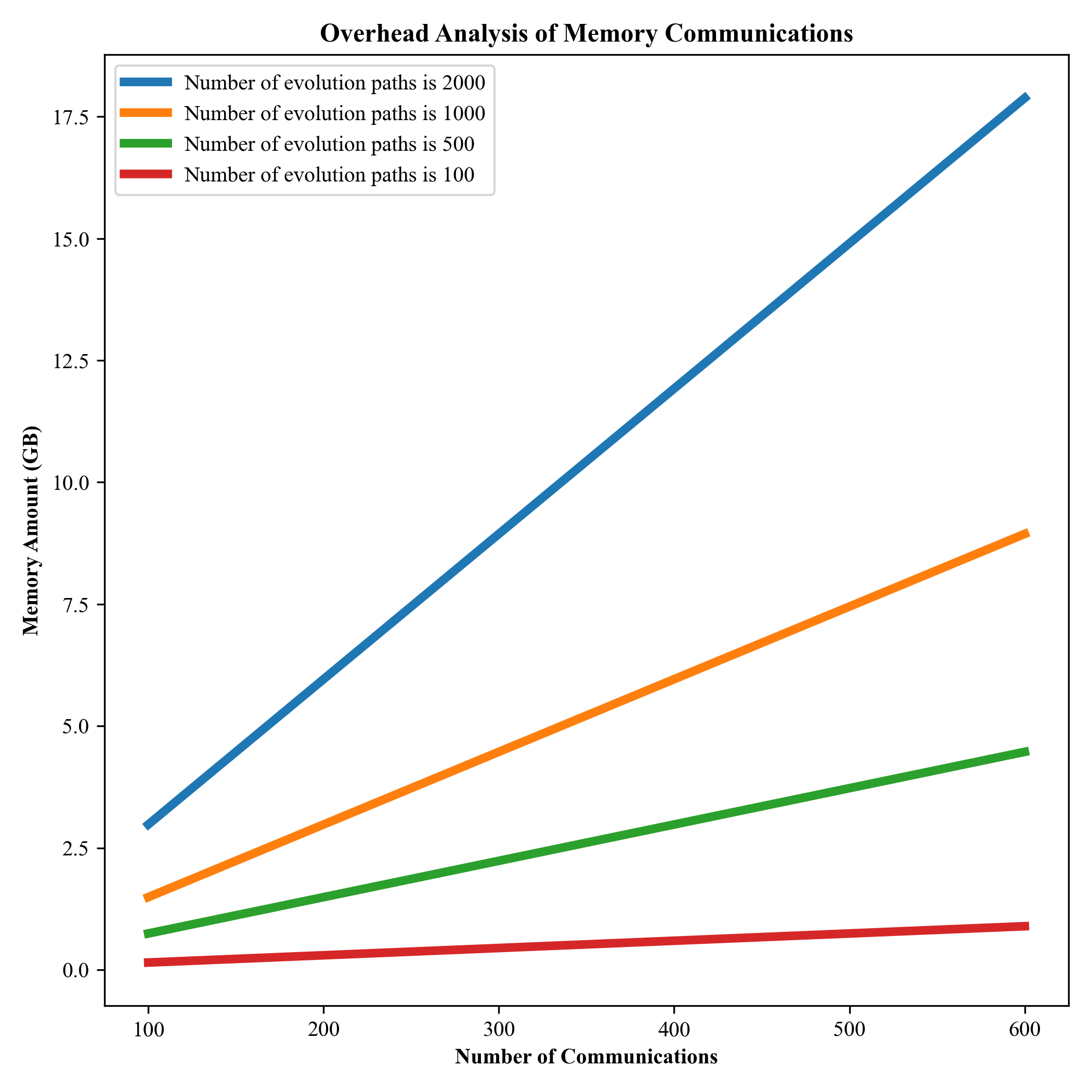}
	\caption{Overhead analysis of memory communications over the network (each time) on a 2000-dimensional fitness function. The $x$-axis is the number of communications (i.e., number of inner-ESs) while the $y$-axis is the memory amount to be communicated over the network.}
	\label{fig:analysis_overhead}
\end{figure}

For our distributed algorithm, a set of evolution paths, distributed over different nodes, are needed to reconstruct the covariance matrix. If this set is too large, it can lead to expensive communication overheads. Otherwise, if this set is too small, it may damage the model richness of the reconstructed covariance matrix. As a result, a trade-off between communication overheads and model richness should be made properly. We can calculate\footnote{\tiny \href{https://github.com/Evolutionary-Intelligence/M-DES/blob/main/figures/plot\_overhead\_of\_memory\_communications.py}{github.com/Evolutionary-Intelligence/M-DES/blob/main/figures/plot\_overhead\_of\_memory\_communications.py}} the amount of memory communications over the network under different settings of number of evolution paths (e.g., 100, 500, 1000, and 2000), as shown in Fig. \ref{fig:analysis_overhead} (on a 2000-dimensional fitness function). Clearly, a high compress ratio (e.g., 100/2000) can significantly reduce the amount of memory communications especially when the level of parallelism is high, while achieving the best performance on most cases (see Figs. \ref{fig:median_unimodal_1}, \ref{fig:median_unimodal_2}, \ref{fig:median_multimodal_1}, and \ref{fig:median_multimodal_2} for empirical demonstrations).

\subsection{Trade-off Analysis of Performance}

In this subsection, we discuss performance trade-offs of our distributed algorithm with one case study called black-box classification from data science (with a non-convex tanh loss function). For this loss function, a business dataset from the popular UCI Machine Learning Repository\footnote{\url{https://doi.org/10.24432/C51G7P}} is employed, leading to an 857-dimensional fitness function (to be minimized).

\begin{figure}
	\centering
	\includegraphics[width=0.48\textwidth, height=0.48\textwidth]{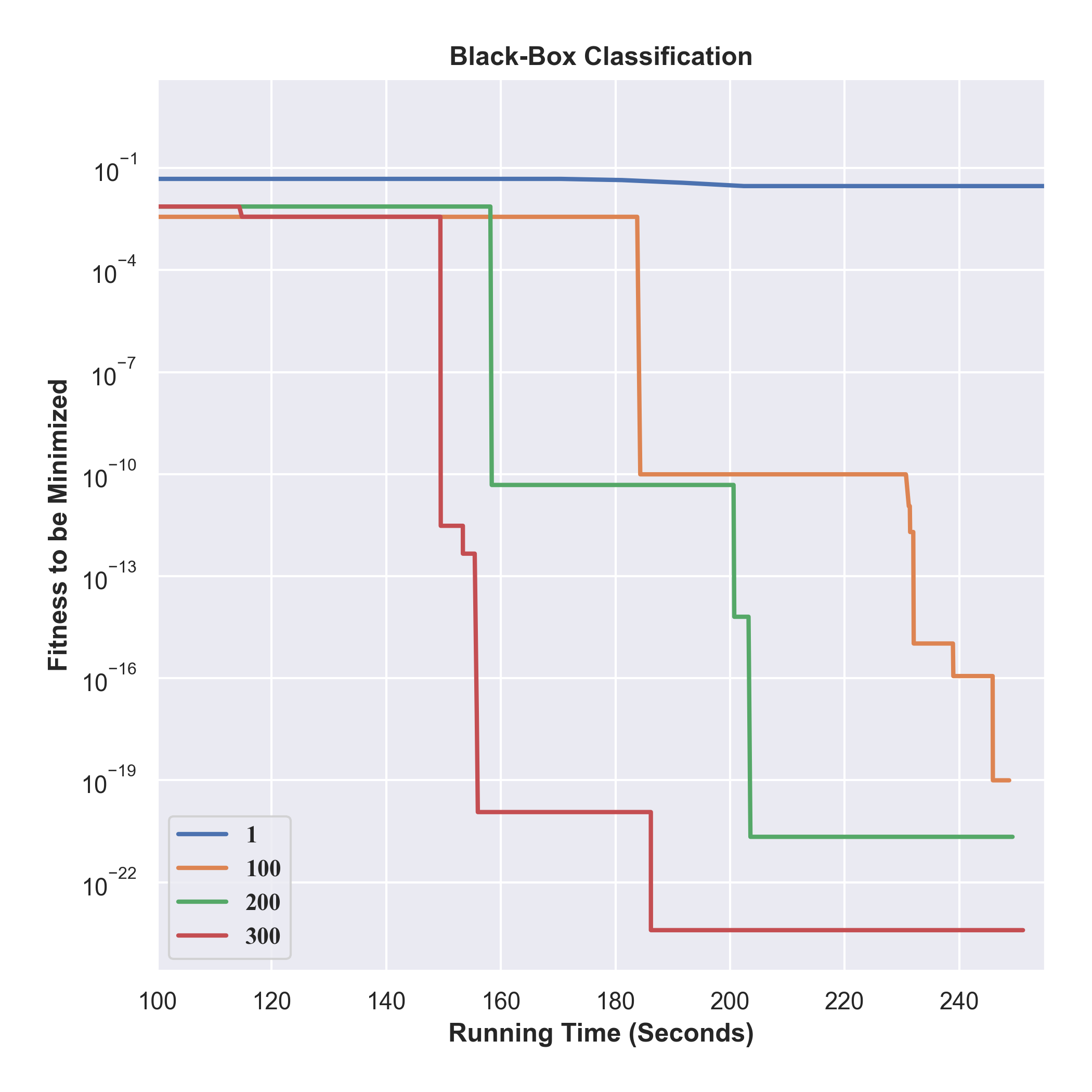}
	\caption{Convergence curves of the best-so-far fitness of our distributed algorithm under four different levels of parallelism (1, 100, 200, and 300).}
	\label{fig:black_box_classification}
\end{figure}

\begin{figure*}[!t]
	\centering
	\subfloat[\scriptsize Number of fitness evaluations (w.r.t. each second) under four different levels (1, 100, 200, and 300) of parallelism.]{\includegraphics[width=0.48\textwidth]{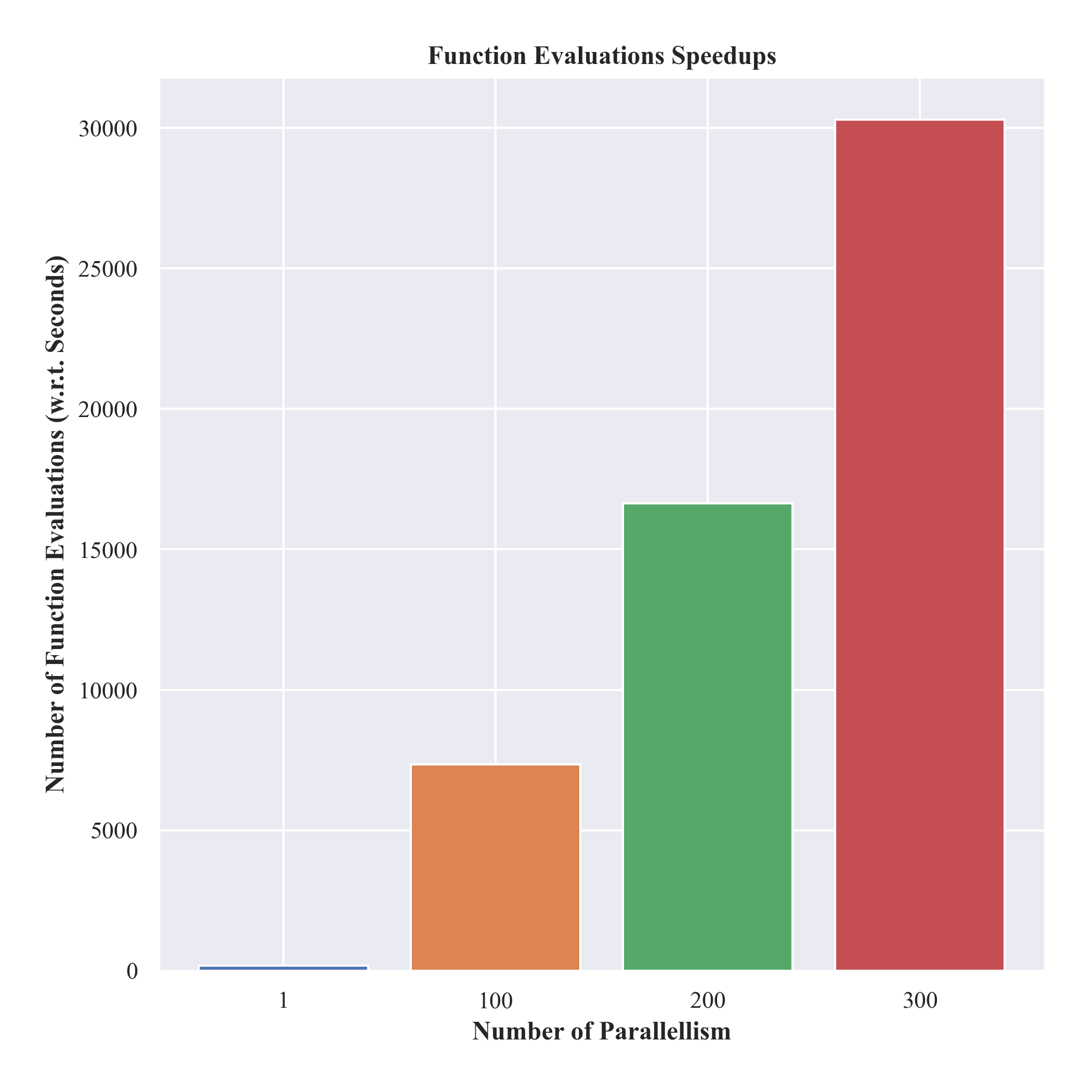}%
		\label{fig:function_evaluations_speedups}}
	\hfil
	\subfloat[\scriptsize Memory amount to be communicated each time under three different settings (25, 440, and 857) of number of evolution paths.]{\includegraphics[width=0.48\textwidth]{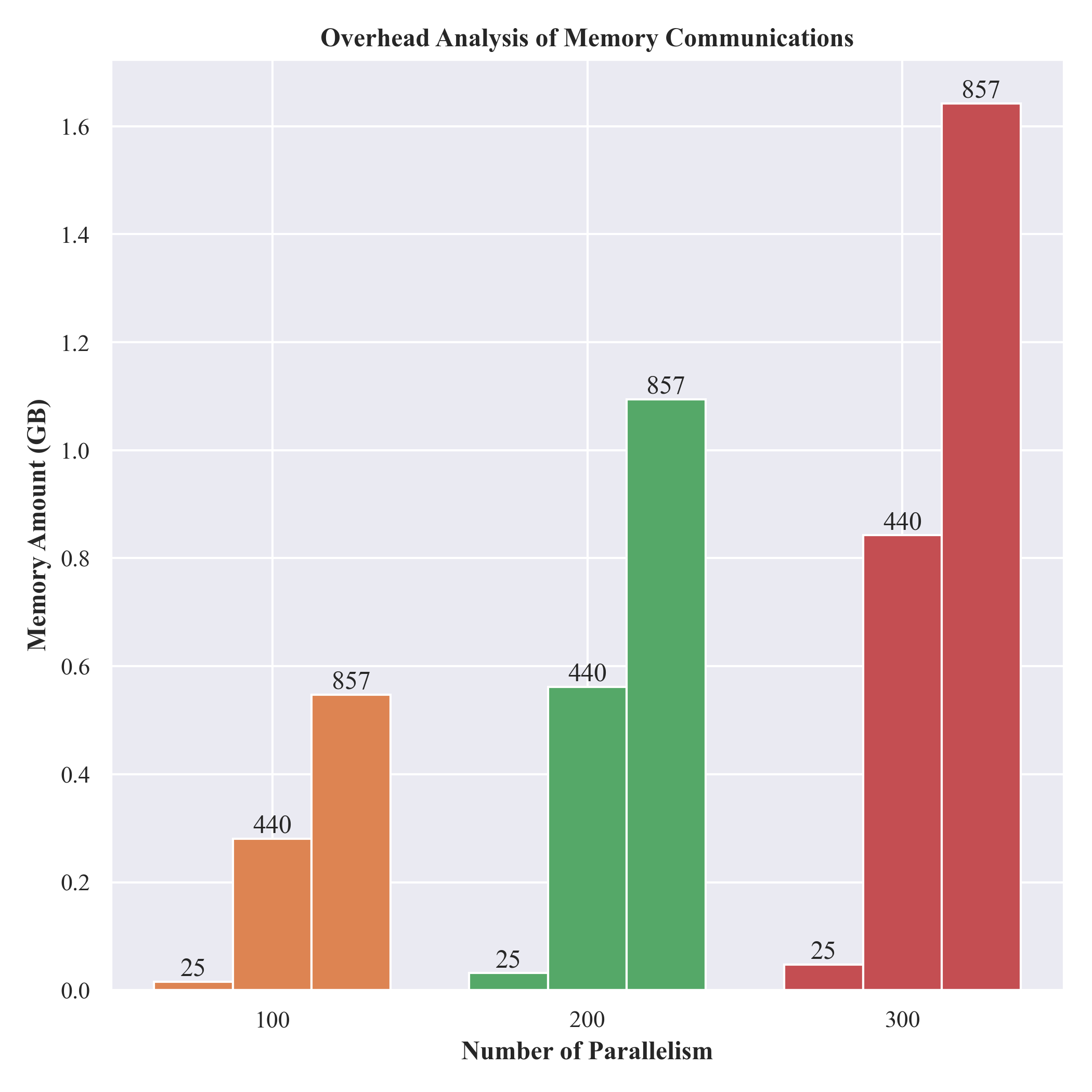}%
		\label{fig:bar_overhead}}

	\caption{Speedups of function evaluations under four different parallelism levels (left) and memory overheads under three different settings of number of evolution paths (right).}
	\label{fig:evaluations_overhead}
\end{figure*}

As we can see from Fig. \ref{fig:black_box_classification} and Fig. \ref{fig:function_evaluations_speedups}, given four different levels of parallelism (that is, 1, 100, 200, and 300), the higher the level of parallelism, the faster (more) the convergence (number of function evaluations). To keep the memory amount to be communicated at a reasonable level, we choose a low number (i.e., 25 in our experiments) as the maximum of evolution paths, as shown in Fig. \ref{fig:bar_overhead}. If the full-ranked covariance matrix was reconstructed (i.e., the number of evolution paths is set to 857), the memory amount to be communicated over the network would increase by $>34$x times, which could result in a significant communication overhead for our distributed algorithm.

\section{Conclusion}

In this paper, we propose a multilevel learning-based meta-framework to parallelize one large-scale variant of CMA-ES called LM-CMA, significantly extending our previous conference paper \cite{10.1007/978-3-031-14721-0_20}. Within this meta-framework, four main design choices are made to control distribution mean update, global step-size adaptation, and CMA reconstruction for effectiveness and efficiency. A large number of comparative experiments show the benefits (and costs) of our proposed meta-framework.

In principle, the proposed distributed meta-framework can be integrated into some other meta-heuristics \cite{aranha2022metaphor} with more or less modifications.

\bibliography{dlmcma}
\bibliographystyle{IEEEtran}

\begin{IEEEbiography}[{\includegraphics[width=1in,height=1.25in,clip,keepaspectratio]{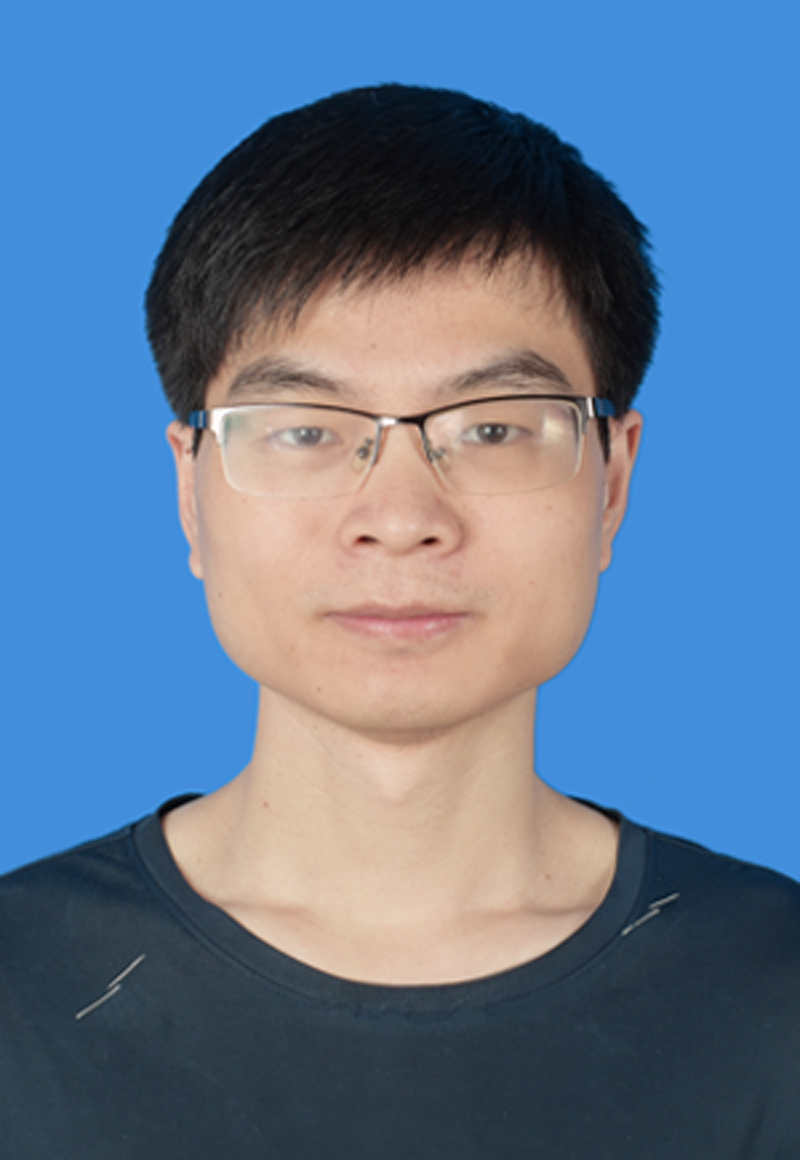}}]{Qiqi Duan}
	is currently pursuing the Ph.D. degree at Harbin Institute of Technology, China while studying in Southern University of Science and Technology, Shenzhen, China. He is one of three core developers of the open-source Python library PyPop7 for evolutionary algorithms and obtained one Best Paper nomination on PPSN-2022. His interests are evolutionary computation, meta-learning, and distributed systems (e.g., swarm intelligence).
\end{IEEEbiography}

\begin{IEEEbiography}[{\includegraphics[width=1in,height=1.25in,clip,keepaspectratio]{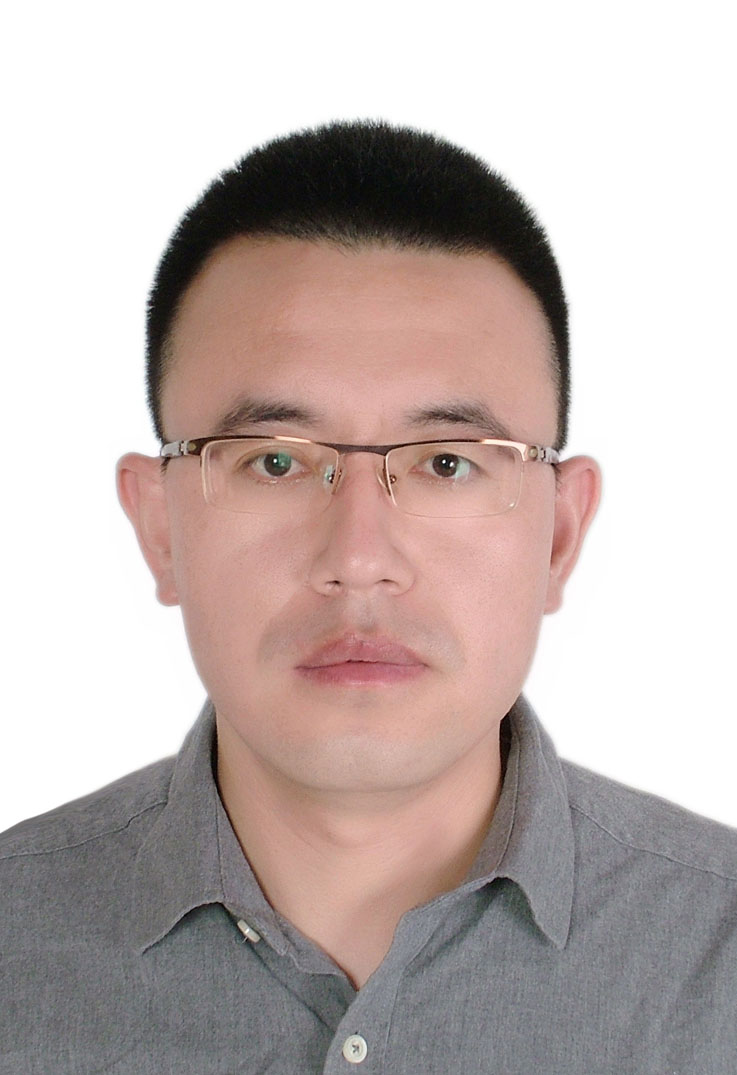}}]{Chang Shao}
	received the BSc degree and MSc degree in Applied Mathematics from Lanzhou University, Lanzhou, China. He is currently pursuing the Ph.D. degree in Computer Science with Australian Artificial Intelligence Institute, University of Technology Sydney, Sydney, NSW, Australia. His research interests include evolutionary computation, swarm intelligence, and dynamic optimization.
\end{IEEEbiography}

\begin{IEEEbiography}[{\includegraphics[width=1in,height=1.25in,clip,keepaspectratio]{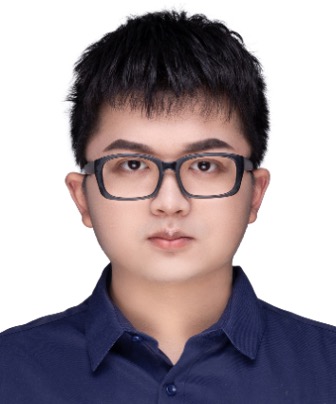}}]{Guochen Zhou}
	received a BSc degree in Computer Science from Chu Kochen Honor College, Zhejiang University, China. He is currently working toward the Master's degree at Southern University of Science and Technology, China. His research interests cover reinforcement learning, offline-to-online fine-tuning, and evolution strategy.
\end{IEEEbiography}

\begin{IEEEbiography}[{\includegraphics[width=1in,height=1.25in,clip,keepaspectratio]{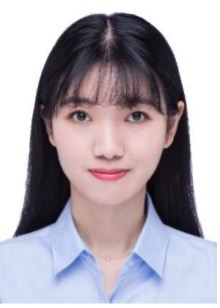}}]{Minghan Zhang}
	received the BSc degree in Mathematics, Optimisation and Statistics and the MSc degree in Statistics from Imperial College London, UK. She is currently working toward the Ph.D. degree with School of Engineering, University of Warwick, Coventry, UK. Her research interests cover evolutionary computation, swarm intelligence and affective computing.
\end{IEEEbiography}

\begin{IEEEbiography}[{\includegraphics[width=1in,height=1.25in,clip,keepaspectratio]{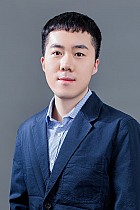}}]{Qi Zhao}
	obtained the Ph.D. degree in Management Science and Engineering from Beijing University of Technology, Beijing, China in 2019 and was a joint Ph.D. student in Computer Science with the University of New South Wales, Canberra, Australia from 2017 to 2018. He is a Research Assistant Professor with the Department of Computer Science and Engineering, Southern University of Science and Technology. His research interests include automated machine learning, operations research, and evolutionary computation.
\end{IEEEbiography}

\begin{IEEEbiography}[{\includegraphics[width=1in,height=1.25in,clip,keepaspectratio]{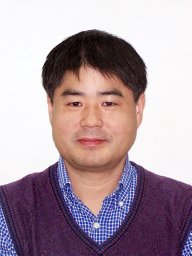}}]{Yuhui Shi}
	(\textit{Fellow}, IEEE) received the Ph.D. degree in electronic engineering from Southeast University, Nanjing, China, in 1992. He is currently a Chair Professor with the Department of Computer Science and Engineering, Southern University of Science and Technology, Shenzhen, China. He has coauthored the book Swarm Intelligence (with Dr. J. Kennedy and Prof. R. Eberhart) and another book Computational Intelligence: Concept to Implementation (with Prof. R. Eberhart). His main research interests are evolutionary computation and swarm intelligence.
\end{IEEEbiography}

\end{document}